\documentclass[conference]{IEEEtran}
\IEEEoverridecommandlockouts
\usepackage{cite}
\usepackage{amsmath,amssymb,amsfonts}
\usepackage{algorithmic}
\usepackage{graphicx}
\usepackage{textcomp}
\usepackage{xcolor}
\def\BibTeX{{\rm B\kern-.05em{\sc i\kern-.025em b}\kern-.08em
    T\kern-.1667em\lower.7ex\hbox{E}\kern-.125emX}}

\usepackage{bm}
\usepackage{multirow}
\usepackage{booktabs}
\usepackage{tabularx}
\usepackage{url}
\usepackage{epsfig}
\usepackage{adjustbox}
\usepackage{rotating}

\DeclareMathOperator*{\argminA}{arg\,min}
    
\begin{document}

\title{An Empirical Study of Parameter Efficient Fine-tuning on Vision-Language Pre-train Model\\
}

\author{\IEEEauthorblockN{Yuxin Tian}
\IEEEauthorblockA{\textit{College of Computer Science,} \\
\textit{Sichuan University}\\
Chengdu, China \\
cs.yuxintian@outlook.com}
\and
\IEEEauthorblockN{Mouxin Yang}
\IEEEauthorblockA{\textit{College of Computer Science,} \\
\textit{Sichuan University}\\
Chengdu, China \\
yangmouxing@gmail.com}
\and
\IEEEauthorblockN{Yunfan Li}
\IEEEauthorblockA{\textit{College of Computer Science,} \\
\textit{Sichuan University}\\
Chengdu, China \\
yunfanli.gm@gmail.com}
\and
\IEEEauthorblockN{Dayiheng Liu}
\IEEEauthorblockA{\textit{College of Computer Science,} \\
\textit{Sichuan University}\\
Chengdu, China \\
losinuris@gmail.com}
\and
\IEEEauthorblockN{Xingzhang Ren}
\IEEEauthorblockA{\textit{School of Software and Microelectronics,} \\
\textit{Peking University}\\
Beijing, China \\
xzhren@pku.edu.cn}
\and
\IEEEauthorblockN{Xi Peng}
\IEEEauthorblockA{\textit{College of Computer Science,} \\
\textit{Sichuan University} and \\
\textit{Engineering Research Center} \\
\textit{of Machine Learning } \\
\textit{and Industry Intelligence,} \\
\textit{Ministry of Education} \\
Chengdu, China \\
pengx.gm@gmail.com}
\and
\IEEEauthorblockN{Jiancheng Lv* \thanks{* Corresponding Author}}
\IEEEauthorblockA{\textit{College of Computer Science,} \\
\textit{Sichuan University} and \\
\textit{Engineering Research Center} \\
\textit{of Machine Learning} \\
\textit{and Industry Intelligence,} \\
\textit{Ministry of Education} \\
Chengdu, China \\
lvjiancheng@scu.edu.cn}
}

\maketitle

\begin{abstract}
Recent studies applied Parameter Efficient Fine-Tuning techniques (PEFTs) to efficiently narrow the performance gap between pre-training and downstream. There are two important factors for various PEFTs, namely, the accessible data size and fine-tunable parameter size. A natural expectation for PEFTs is that \textit{the performance of various PEFTs is positively related to the data size and fine-tunable parameter size.} However, according to the evaluation of five PEFTs on two downstream vision-language (VL) tasks, we find that such an intuition holds only if the downstream data and task are not consistent with pre-training. For downstream fine-tuning consistent with pre-training, data size no longer affects the performance, while the influence of fine-tunable parameter size is not monotonous. We believe such an observation could guide the choice of training strategy for various PEFTs.
\end{abstract}


\section{Introduction}
Vision-language pre-training (VLP) has emerged as a fundamental paradigm to boost the performance of downstream VL tasks.
Most existing works boost the performance of the pre-trained model by designing novel pre-training tasks~\cite{wangOFAUnifyingArchitectures2022}, increasing the size of both pre-training dataset~\cite{radfordLearningTransferableVisual2021} and the model parameters~\cite{liAlignFuseVision2021, liBlipBootstrappingLanguageimage2022}.
Although VLP has shown promising zero-shot performance on the downstream tasks, fine-tuning still plays an indispensable role in narrowing the gap between pre-training and downstream domains.

Taking the fine-tuning cost into consideration, numerous Parameter Efficient Fine-Tuning (PEFT) methods have been proposed to adapt the VLP models to downstream tasks in an efficient manner.
More specifically, PEFT methods only fine-tune a few parameters while freezing the most pre-trained parameters.
In this study, we mainly focus on the PEFT methods that insert additional parameters into different positions of the VLP models, which we refer to as the exogenous PEFT.
For clarity, we divide the exogenous PEFT methods into \textit{embedding composition} ones (\textit{e.g.,} prompt-tuning~\cite{lesterPowerScaleParameterefficient2021} and prefix-tuning~\cite{liPrefixTuningOptimizingContinuous2021}) and \textit{layer composition} ones (\textit{e.g.,} Adapter~\cite{houlsbyParameterefficientTransferLearning2019,heUnifiedViewParameterEfficient2022} and LoRA~\cite{huLoRALowRankAdaptation2022}), according to the inserted position. 
Almost all existing PEFT methods seek to achieve competitive performance compared to the full fine-tuning counterparts while embracing high training efficiency.
Intuitively, one may expect that the performance of various PEFTs is positively related to the accessible data size and fine-tunable parameter size.

\begin{figure}[t]
\centering
\epsfig{figure=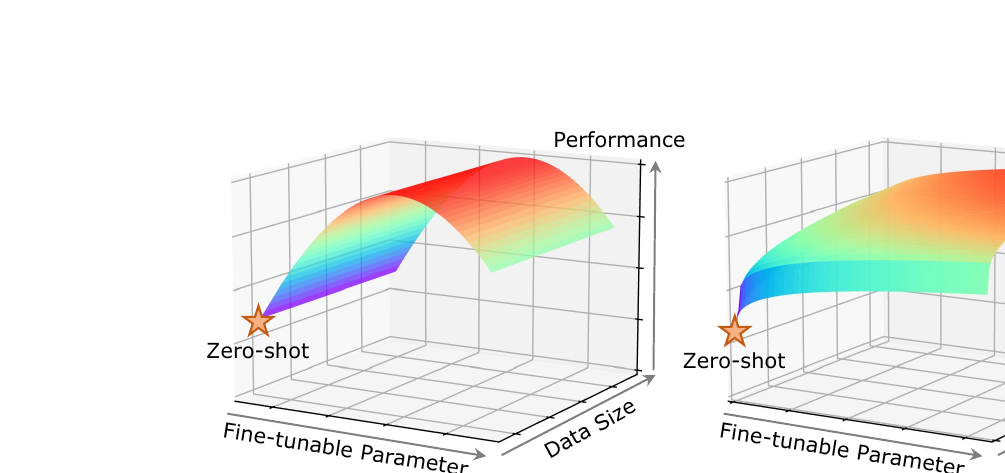,width=8.5cm}
\caption{The performance is only affected by the size of fine-tunable parameters when the downstream task and data are consistent with pre-training. Otherwise, the performance is positively related to data and parameter size.}
\label{fig:f2}
\end{figure}

To evaluate such an expectation, we conduct an empirical analysis with five PEFTs on two downstream VL tasks. 
Specifically, we first offer a novel unified view of the investigated prompt-tuning, prefix-tuning, LoRA, serial adapter-tuning, and parallel adapter-tuning.
Then, considering the differences between the pre-training and downstream fine-tuning, the accessible downstream data size, and the size of the fine-tunable parameters, we conduct a series of experimental evaluations on two widely-used VL datasets, \textit{i.e.,} MSCOCO Caption and VQAv2.
It is worth noting that the image caption task is usually adopted as a pre-training task in the VLP, but not with VQA.
Therefore, the image caption and the corresponding data in the downstream tasks could be regarded as consistent with the VLP, while the VQA one is inconsistent.

The contributions of this study mainly lie in the new empirical observations.
As illustrated by Fig.~\ref{fig:f2}, if the downstream task and dataset are not consistent with pre-training, the data size and the fine-tunable parameter size are positively related to the performance.
If consistent, the data size no longer affects the performance of various PEFTs while the influence of fine-tunable parameter size is not monotonous.
We believe that such observations would guide the training strategy design of various PEFTs.
Furthermore, our experimental results also reveal an additional phenomenon: Considering the training efficiency and performance, layer composition (\textit{e.g.,} LoRA) could be a better choice for the downstream adaptation of the VLP model.

\section{Related Work}
\label{sec:related}

\subsection{PEFT for VLP model}
A growing body of research has been devoted to finding parameter-efficient alternatives to adapt large-scale VLP models to downstream tasks and to reduce the cost of various aspects such as memory and storage. 
The representative works for the first line are prompt-tuning~\cite{lesterPowerScaleParameterefficient2021} and prefix-tuning~\cite{liPrefixTuningOptimizingContinuous2021}.
The early prompt-based works employ prompt-tuning~\cite{lesterPowerScaleParameterefficient2021} to manage the few-shot visual tasks~\cite{zhouLearningPromptVisionLanguage2022}.
The most recent researches expand the application of prefix-tuning into the VLP~\cite{khattakMapleMultimodalPrompt2023} model and achieve comparable performance with full finetuning at low cost.
The other line of work is based on the adapter~\cite{houlsbyParameterefficientTransferLearning2019} and LoRA~\cite{huLoRALowRankAdaptation2022}.
Most recent works~\cite{zhangLLaMAAdapterEfficientFinetuning2023, hu2023vl} resort to the adapter layer to extend the multimodal abilities of the generative large-scale language model.
Besides, LoRA~\cite{huLoRALowRankAdaptation2022} are also used to adapt text-only LLM for multimodal tasks~\cite{yeMPLUGOwlModularizationEmpowers2023}.
Although all of these methods have demonstrated their effectiveness, there is currently no analysis to investigate how the data size and fine-tunable parameter size affect various PEFTs.

\subsection{Empirical PEFT analysis}
Recent studies have conducted numerous experiments to examine the possible factors that influence the performance and robustness of PEFT. 
Chen et.al~\cite{chenRevisitingParameterEfficientTuning2022} reveal that full fine-tuning cannot be entirely replaced by PEFT approaches currently in NLP since it cannot attain superior performance to full fine-tuning when given adequate fine-tuning budget and data size.
Sung et.al~\cite{sungVLADAPTERParameterEfficientTransfer2022} finds that the vanilla adapter~\cite{houlsbyParameterefficientTransferLearning2019} achieved the best VL task performance among its variants.
Regarding robustness, Chen et.al~\cite{chenBenchmarkingRobustnessAdaptation2023a} deem that neither full fine-tuning nor PEFT approaches consistently provide robustness for data corruption.
Differently, this study investigates an inevitable problem \textit{Are the data size and fine-tunable parameter size positively related to the VL performance of various PEFTs?}

\section{A Unified View for PEFT}

\begin{figure}[t]
\centering
\includegraphics[width=8.5cm]{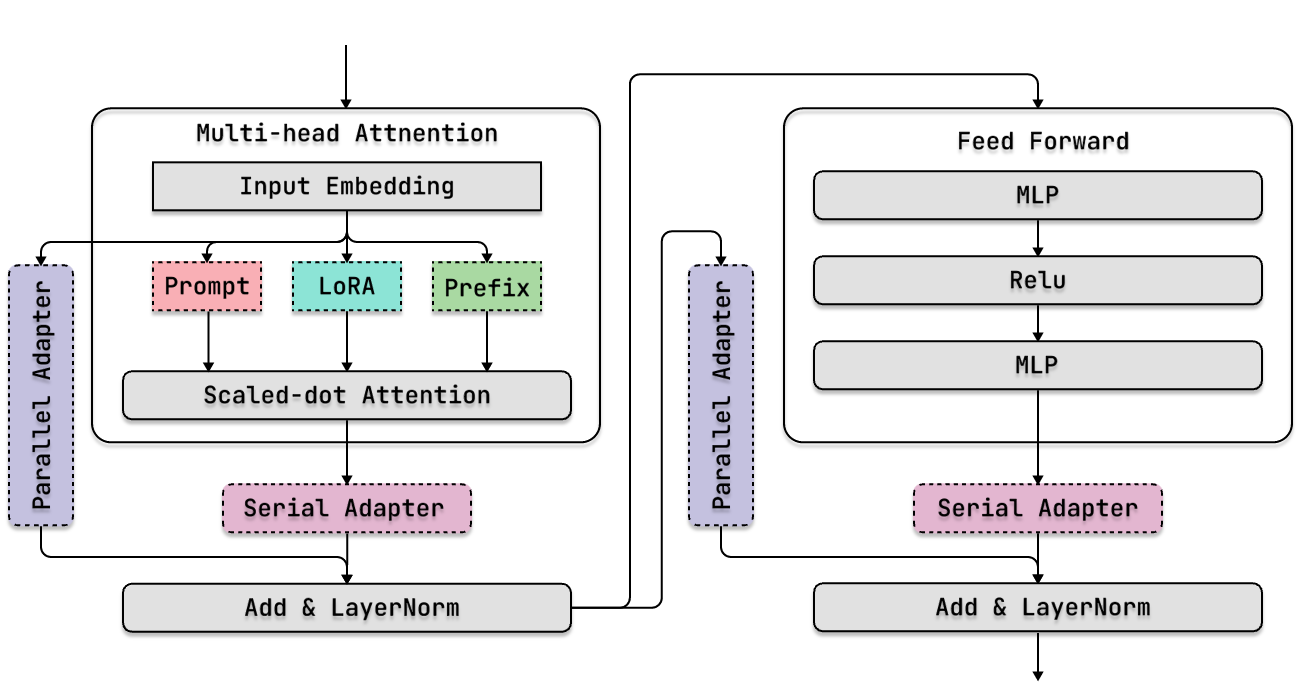}
\caption{Unified view of evaluated PEFT methods within a transformer block.}
\label{fig:peft-unified}
\end{figure}

In this section, we briefly review the investigated PEFTs from a unified perspective view.
The exogenous parameter fine-tuning efficiently trains the VLP model by incorporating additional trainable parameters into the input, \textit{e.g.,} \textbf{prompt-tuning} and \textbf{prefix-tuning}, or intermediate layers of the model, \textit{e.g.,} \textbf{LoRA, serial adapter-tuning}, and \textbf{parallel adapter-tuning}.
To be specific, given a pre-trained VLP model $F$ with parameters $\bm{\Theta}$, PEFT resorts to fine-tuning the additional parameters $\mathbf{\Phi}$ to adapt the pre-trained model $F$ from pre-training task to downstream task, \textit{e.g.,} VQA. 
In formal, with the input embedding $\mathbf{x} \in \mathbb{R}^{n \times d}$ for a transformer block and a ground truth $\mathbf{y}$, the objective of such PEFT could be unified as follow:
\begin{equation}
\label{eq1}
\begin{aligned}
	\argminA_{\bm{\Phi}} & \ \mathcal{L}\left(F\left(\mathbf{x};\bm{\Theta},\bm{\Phi}\right),\mathbf{y}\right) \\
	= & \ \mathcal{L}\left(f_n\circ \cdots \circ f_1(\mathbf{x};\bm{\Theta}_1,\bm{\Phi}_1),\mathbf{y}\right),
\end{aligned}
\end{equation}
where $\mathcal{L}$ and $f_i$ denotes the loss function of downstream task and the $i$-th transformer block, respectively.
According to the position of the extra parameters introduced by PEFT, we categorize the tested prompt-tuning and prefix-tuning as \textit{embedding composition} and serial adapter-tuning, parallel adapter-tuning, and LoRA as \textit{layer composition}.
We will elaborate on them in the following. 

\subsection{Full finetuning}
The most straightforward way to adapt the pre-trained VLP models to downstream tasks is full finetuning. 
Yet, directly updating the full set of $\bm{\Theta}$ costs vast computational resources, particularly as model size continues to increase. 

\subsection{Embedding composition}
Prior study~\cite{lesterPowerScaleParameterefficient2021} proposed ``continuous'' prompt which concatenates trainable parameters $\bm{\Phi}$ to the embedding of the transformer layers and hence we entitle them as embedding composition. 
For example, prompt-tuning~\cite{lesterPowerScaleParameterefficient2021} prepends the model input embedding $\mathbf{x} \in \mathbb{R}^{n \times d}$ with a learnable embedding $\bm{\Phi}:= \mathbf{p} \in \mathbb{R}^{p \times d}$ optimized directly through gradient descent, where $n$/$p$ denotes the length of the input/prompt embedding and $d$ is the dimension of them. 
According to Eq.\ref{eq1}, the objective of prompt-tuning could be rewritten following: 
\begin{equation}
	\argminA_{\mathbf{p}} \mathcal{L}\left(F\left(\left[\mathbf{p}:\mathbf{x}\right]\right),\mathbf{y}\right),
\end{equation}
where $[\cdot:\cdot]$ indicates the concatenation operation. 
Instead of adding parameters to the input embedding of the model, prefix-tuning~\cite{liPrefixTuningOptimizingContinuous2021} prepends trainable tokens, \textit{i.e.,} $\bm{\Phi}:=\{\mathbf{p}_i\}^{n}$, to the hidden states of all the transformer blocks. 
Formally, the corresponding objective is as follows:
\begin{equation}
	\argminA_{\{\mathbf{p}_i\}^{n}_1} \mathcal{L}\left(f_n^{p} \circ f^{p}_{n-1} \circ \cdots \circ f_1^{p}(\mathbf{x}),\mathbf{y}\right),
\end{equation}
where $f_i^{p}(\mathbf{x}):=f_i(\left[\mathbf{p}_i:\mathbf{x}\right])$. 
Usually, one could use an additional MLP encoder $P$ to encode $\{\mathbf{p}_i\}^n$ for better training stability. 

\subsection{Layer composition}
Differing from embedding composition, layer composition inserts an extra sub-network with few trainable parameters into transformer blocks. 
For example, \cite{houlsbyParameterefficientTransferLearning2019} firstly proposes a vanilla adapter layer that consists of two fully connected layers.
Following effort~\cite{heUnifiedViewParameterEfficient2022} extends it and proposes a parallel variant instead of inserting the adapter layer sequentially.  
Formally, inserting two serial/parallel adapter layers $\{A_{i,1}^{\{s,p\}},A_{i,2}^{\{s,p\}}\}$ to the $i$-th transformer block as additional parameters $\bm{\Phi}:=\{A_{i,1}^{\{s,p\}},A_{i,2}^{\{s,p\}}\}^{n}$, one could have the following objective of adapter-tuning:
\begin{equation}
	\argminA_{\{A_{i,1},A_{i,2}\}^{n}} \mathcal{L}\left(f_n^{a} \circ f^{a}_{n-1} \circ \cdots \circ f_1^{a}(\mathbf{x}),\mathbf{y}\right),
\end{equation}
where $f^{a}_i$ could be the $i$-th transformer block with either sequential adapter layer $f_i^{sa}$ or parallel adapte layer $f_i^{pa}$:
\begin{equation}
	\begin{aligned}
		f_i^{sa}(\mathbf{x}):= & AL \circ A^{s}_{i,2} \circ FFN \circ \\ 
		& AL \circ A^{s}_{i,1} \circ MHA(\mathbf{x},\mathbf{x}),
	\end{aligned}
\end{equation}
\begin{equation}
		f_i^{pa}(\mathbf{x}) := AL (A^{p}_{i,2}(h_{med}) + FFN(h_{med}))
\end{equation}
\begin{equation}
    h_{med} = AL\left(A^{p}_{i,1}(\mathbf{x}) + MHA(\mathbf{x},\mathbf{x})\right)
\end{equation}
\begin{equation}
	A^{\{s,p\}}_i(\mathbf{x}) := ReLU(\mathbf{x}\mathbf{W_{down}})\mathbf{W_{up}} + \mathbf{x}
\end{equation}
where $AL$, $FFN$, and $MHA$ denote the skip-connection and layernorm, feed-forward, and multi-head attention within a transformer block, respectively. 
Further, the matrices, \textit{i.e.,} $\mathbf{W_{down}}\in \mathbb{R}^{d_h \times l}$ and $\mathbf{W_{up}}\in \mathbb{R}^{l \times d_h}$, are the trainable parameters of a adapter layer, where $d_h$ denotes the dimension of hidden state.

In addition to the adapter-tuning, LoRA~\cite{huLoRALowRankAdaptation2022} inserts reparametrized fully connected layers into the self-attention layers, which is inspired by the intrinsic dimensionality~\cite{aghajanyanIntrinsicDimensionalityExplains2021}. 
In detail, LoRA utilizes the low-rank decomposition matrices~\cite{aghajanyanIntrinsicDimensionalityExplains2021} to reparameterize the additional MLP layers. 
In formal, given the number of head $k$, the hidden state of $i$-th attention head $\mathbf{h}_i$, and the trainable parameters, \textit{i.e.,} $\bm{\Phi}:=\{\mathbf{\Phi}_{i,1},\mathbf{\Phi}_{i,2}\}^{n}$ which are plugged into the $MHA$ layers, one could have the following objective for LoRA:
\begin{equation}
	\argminA_{\{\mathbf{\Phi}_{i,1},\mathbf{\Phi}_{i,2}\}^{n}} \mathcal{L}\left(f_n^{l} \circ f^{l}_{n-1} \circ \cdots \circ f_1^{l}(\mathbf{x}),\mathbf{y}\right).
\end{equation}
The transformer block with LoRA layer could be as follows:
\begin{equation}
	\begin{aligned}
		f_i^{l}(\mathbf{x}):= & AL \circ FFN \circ AL \circ MHA^{l}_i(\mathbf{x},\mathbf{x}),
	\end{aligned}
\end{equation}
\begin{equation}
	MHA^{l}_i(\mathbf{g},\mathbf{x})=[\mathbf{h}_i^1:\mathbf{h}_i^2:\cdots:\mathbf{h}_i^k]\mathbf{W}_{o},
\end{equation}
\begin{equation}
	\begin{aligned}
		\mathbf{h}_i^j= & \ Attn\left.(\mathbf{x}\left(\mathbf{W}^j_q+\mathbf{\Phi}_{i,1}\right)\right., \\
		& \left. \mathbf{g}\left.(\mathbf{W}^j_k+\mathbf{\Phi}_{i,2}\right.), \mathbf{g}\mathbf{W}^j_v \right.),
	\end{aligned}
\end{equation}
\begin{equation}
	\mathbf{\Phi}_{i,m}=\mathbf{B}\mathbf{A}, 
\end{equation}
where the parameters of multi-head attention are
\begin{equation}
	\mathbf{W}_o \in \mathbb{R}^{d \times d}, \mathbf{W}_q^j, \mathbf{W}_k^j, \mathbf{W}_v^j  \in \mathbb{R}^{d \times \frac{d}{k}},
\end{equation}
and the low-rank trainable matrices for the LoRA layer are
\begin{equation}
	\mathbf{B} \in \mathbb{R}^{d \times r}, \mathbf{A} \in \mathbb{R}^{r \times \frac{d}{k}}, r \ll \frac{d}{k}.
\end{equation}
$MHA$ functions as self-attention, such that $\mathbf{g}=\mathbf{x}$.
Otherwise, it operates as cross-attention.

\begin{figure*}[t]
\centering
\epsfig{figure=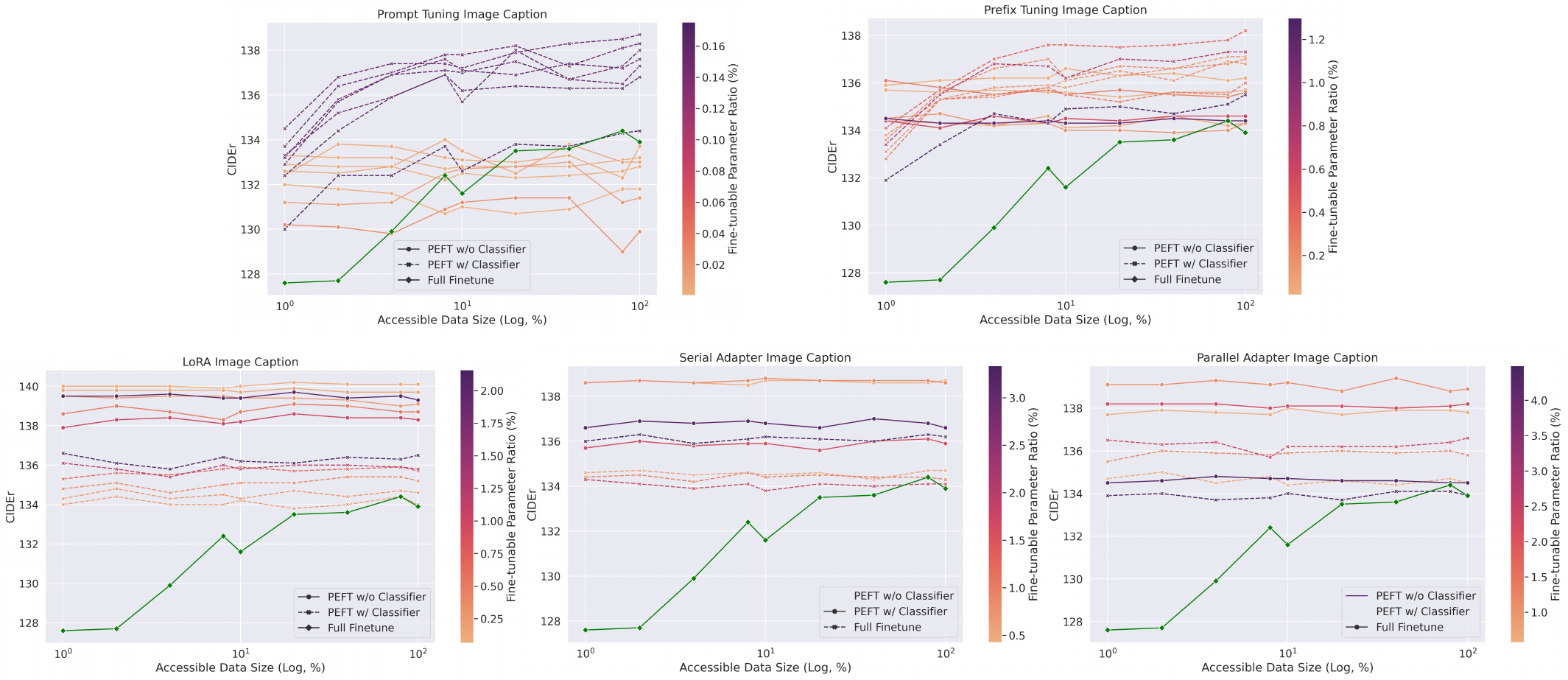,width=17cm}
\caption{\textbf{Layer composition PEFTs achieve better performance than embedding composition on MSCOCO Caption.} Both layer and embedding composition PEFTs could achieve comparable performance with full fine-tuning. The performance of the tested PEFTs is regardless of the accessible data size. Increasing the size of fine-tunable parameters by simultaneously fine-tuning the final classifier only improves the performance of prompt-tuning, and hurts that of the layer composition.}
\label{fig:ic-data}
\end{figure*}

\begin{figure*}[t]
\centering
\epsfig{figure=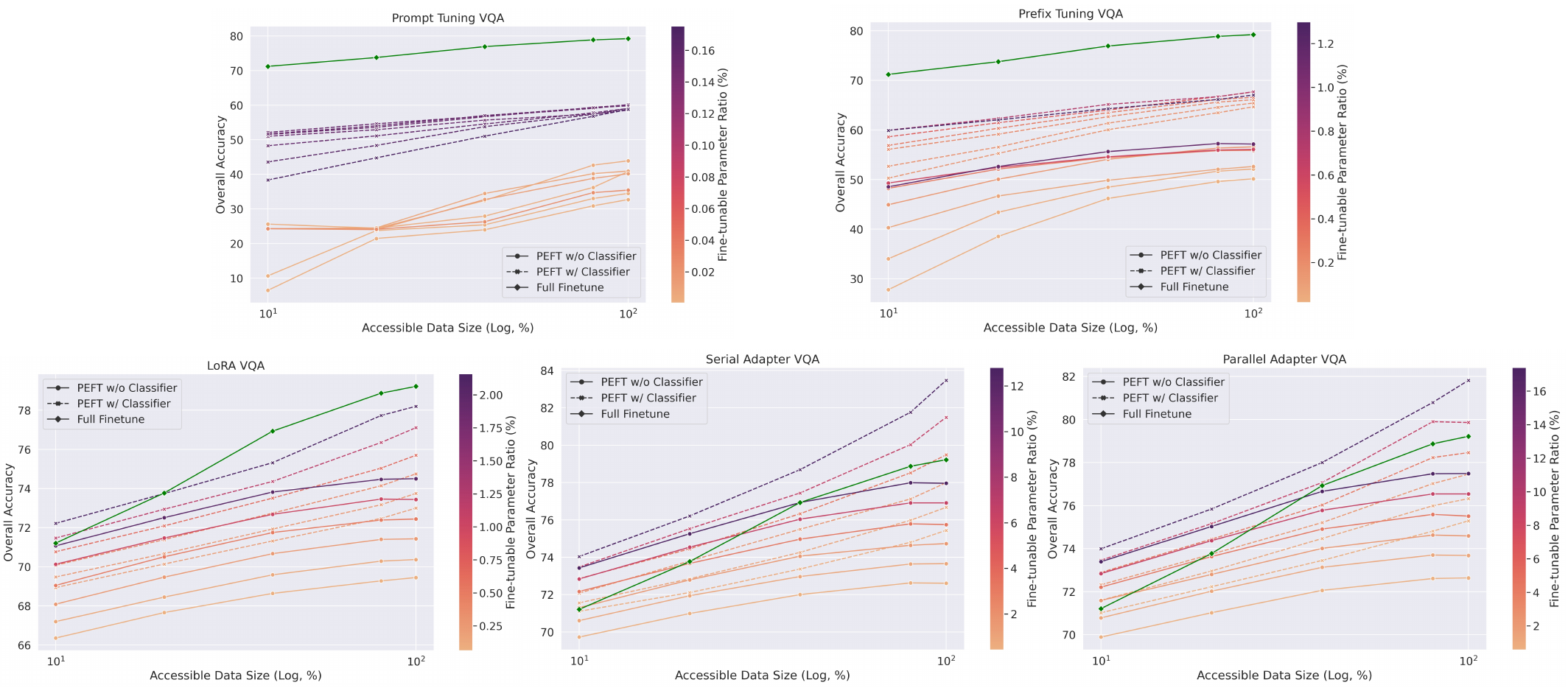,width=17cm}
\caption{\textbf{Layer composition PEFTs achieve better performance than embedding composition on VQAv2.} Empirically, Layer composition PEFTs could achieve comparable performance with full fine-tuning, while embedding composition PEFTs cannot. The performance of the tested PEFTs is positively correlated to the accessible training data and fine-tunable parameters. Additionally, simultaneously fine-tuning the final classifier of the model could further boost the performance and even achieve superior performance than full fine-tuning.}
\label{fig:vqa-data}
\end{figure*}

\section{Experimental setup}
\label{sec:exp-setup}
\subsection{Base model}
We adopt the recently-proposed VLP model, namely, mPLUG~\cite{liMPLUGEffectiveEfficient2022}, as our base model, which employs the discriminative-generative pre-training tasks~\cite{liAlignFuseVision2021,liBlipBootstrappingLanguageimage2022}. 
In brief, mPLUG adopts the pre-trained CLIP-ViT~\cite{radfordLearningTransferableVisual2021} (ViT-B/16) as visual encoder and two $BERT_{base}$~\cite{devlinBERTPretrainingDeep2019} models for textual-visual feature fusing and text decoding (See supplementary material for more details).

\subsection{Dataset and task}
We investigate two widely-used and distinct VL down-streaming tasks in this study, \textit{i.e.,} visual question answering on VQAv2~\cite{goyalMakingVQAMatter2019} and image captioning on MSCOCO Caption~\cite{chenMicrosoftCOCOCaptions2015} (See supplementary material for dataset statistics).
The VQA is a multi-modal image understanding task that requires the VLP model to answer the textual questions referring to the corresponding image. 
Following~\cite{liAlignFuseVision2021,liBlipBootstrappingLanguageimage2022}, we treat it as an answer generation without constraints for better generality. 
Image captioning is a multi-modal text generation task that asks for a VLP model to generate an accurate and fluent caption for a given image.

\subsection{Implementation details}
Both for captioning and VQA, the visual encoder (CLIP-ViT-B/16) takes the resized 256 $\times$ 256 image as input.
The text encoder takes a caption prompt \{\textit{a picture of}\} and a  question \{\textit{Question: \{\#question\} Answer: }\} as the input for captioning and VQA, respectively.
Then the caption and answer would be generated by the following text decoder.
During training, we used the AdamW optimizer with a weight decay of 0.05.
The learning rate is warmed up to the highest learning rate in the first epoch, and decayed to the lowest learning rate following a cosine schedule.
All the experiments are optimized by cross-entropy loss and conducted on four V100-32G-SMX2 GPUs with a total batch size of 256 (See supplementary material for details).
We report the best performance metric amongst the whole training.
To be specific, the CIDEr~\cite{vedantamCIDErConsensusbasedImage2015} on Karpathy-test split is used for image captioning, whilst the overall accuracy on VQAv2 validation split is reported~\footnote{The submission on the test set is limited: \url{https://eval.ai/web/challenges/challenge-page/830/phases}} by following~\cite{liAlignFuseVision2021,liMPLUGEffectiveEfficient2022}.

\subsection{Evaluation protocol}
We conduct experiments on two VL downstream tasks with five PEFTs, \textit{i.e.,} prompt-tuning, prefix-tuning, LoRA, serial adapter-tuning, and parallel adapter-tuning, to investigate \textit{how the accessible data size and the fine-tunable parameter size affects the performance of them.}
Firstly, the available data for different tasks varies, so we employ a random sampling to select diverse proportional Image-Caption and QA pairs from MSCOCO Caption and VQAv2 datasets for training.
Second, the available hardware constraints on the available fine-tunable parameters of various PEFT.
To explore \textit{how the size of the fine-tunable parameters influence the performance of different PEFTs}, we modulate the length of the prompt, the length of the prefix, the rank of LoRA, as well as the hidden size of both the serial adapter and the parallel adapter. (See supplementary material for the details).

\section{Results and Analysis}

\subsection{PEFT with various data sizes}
In this section, we investigate how the accessible data size influences the performance of various PEFTs.
According to Fig.~\ref{fig:ic-data}, one could find that the performance of the tested PEFTs is regardless of different accessible data sizes of MSCOCO Caption, which is different from our intuition.
However, from Fig.~\ref{fig:vqa-data}, the performance of all the tested PEFTs steadily increases when the accessible training data size of VQAv2 increases.
This could be imputed to that captioning on MSCOCO Caption is consistent with pre-training while VQAv2 is not.
To be specific, vision-language pre-training usually utilizes a large-scale of image-text pairs~\cite{liMPLUGEffectiveEfficient2022,liAlignFuseVision2021,liBlipBootstrappingLanguageimage2022} and the downstream data may inevitably be used for pre-training, \textit{e.g.,} MSCOCO Caption.

\subsection{PEFT with various parameter sizes}
In this section, we investigate how the fine-tunable parameter size affects the performance of various PEFTs.
From Fig.~\ref{fig:vqa-data} and~\ref{fig:ic-data}, the fine-tunable parameter size does affect the performance of various PEFTs.
To be specific, the fine-tunable parameter size is positively related to the performance when one adopts the VL downstream task and data that are different from pre-training, \textit{e.g.,} VQA.
Such a phenomenon is consistent with our intuition. 
If we increase the fine-tunable parameter size by simultaneously fine-tuning the final classifier with various PEFTs, the performance on VQAv2 can be further improved and they could even achieve superior performance than full fine-tuning.

However, when the VL downstream task and data are consistent with the pre-training, \textit{i.e.,} captioning on MSCOCO Caption, there exists an optimal fine-tunable parameter choice, which is different from our intuition.
Additionally, increasing the fine-tunable parameter size by simultaneously fine-tuning the final classifier could not always improve the performance on MSCOCO Caption.
We deem such a phenomenon also attributes to the consistency between downstream fine-tuning and pre-training.
When the downstream fine-tuning is consistent with pre-training, too many fine-tunable parameters could lead to over-fitting. (See supplementary material for further discussion.)

\subsection{Further analysis}
\begin{figure}[t]
\centering
\epsfig{figure=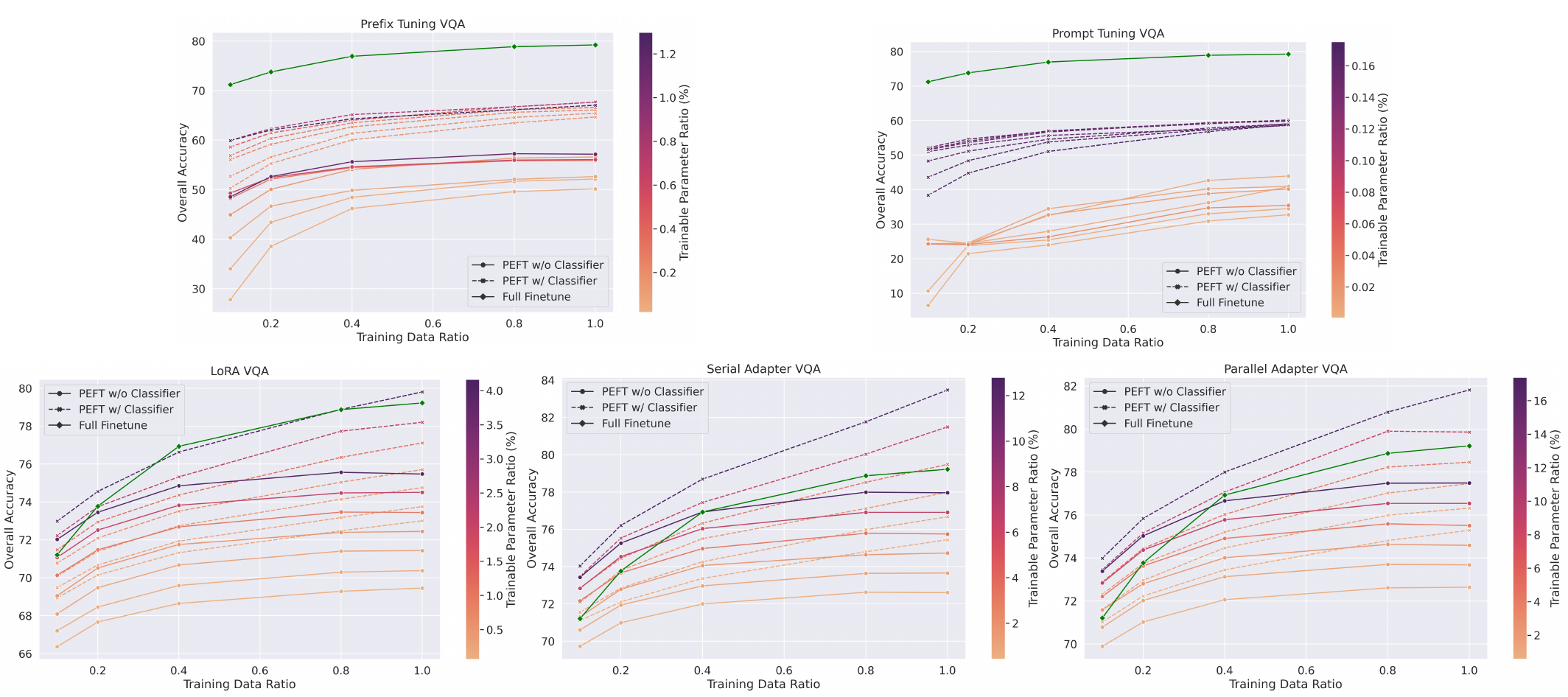,width=8.5cm}
\caption{The comparison of various PEFTs.}
\label{fig:efficient}
\end{figure}

Besides, our experimental results also provide a comparison of various PEFTs.
Fig.~\ref{fig:ic-data} indicates that the tested PEFTs always outperform the full fine-tuning except for the prompt-tuning on MSCOCO Caption.
Such phenomenon may be attributed to the fact that prompt-tuning possesses less than 0.16\% fine-tunable parameters which is less than others.
On the other hand, the embedding composition PEFTs lag far from full fine-tuning, while the layer embedding composition PEFTs achieve comparable performance to the full fine-tuning, illustrated by Fig.~\ref{fig:vqa-data}.
Eventually, by considering the product of data size and fine-tunable parameter size as computation cost, one could easily find which PEFT is better for the tested VLP model.
According to Fig.~\ref{fig:efficient}, the layer composition offers superior training efficiency and performance on the two downstream VL tasks.

\section{Conclusion}
In this paper, we conduct a comprehensive study on five PEFTs on two VL downstream datasets to investigate: \textit{how accessible data size and fine-tunable parameter size affect the performance of various PEFTs.} 
Referring to our experiments, we find that if the downstream data and task are not consistent with pre-training, increasing the fine-tunable parameter size or accessible data size benefits the performance of PEFT as expected.
Nevertheless, if consistent, the data size does not affect the performance and the fine-tunable parameter holds an optimal size choice. 
Such a phenomenon could guide the training strategy design of various PEFTs.
We leave the exploration of the influence of the base model, more downstream tasks, and data to future work.

\clearpage

\bibliographystyle{IEEEbib}
\bibliography{main}

\clearpage

\appendices

\section{Experimental Setup}
\subsection{Data Details}
We investigate two widely-used and distinct VL down-streaming tasks in this study, \textit{i.e.,} visual question answering on VQAv2~\cite{goyalMakingVQAMatter2019} and image captioning on MSCOCO Caption~\cite{chenMicrosoftCOCOCaptions2015}.
The dataset statistics are shown in Table~\ref{tab:dataset}.

\subsection{Base model}
We utilize the recent VLP model, namely, mPLUG~\cite{liMPLUGEffectiveEfficient2022}, as our base model, which follows the discriminative-generative pre-train tasks~\cite{liAlignFuseVision2021,liBlipBootstrappingLanguageimage2022}. 
In general, mPLUG resorts to three discriminative tasks, \textit{i.e.,} Image-Text Contrastive Learning~\cite{liAlignFuseVision2021}, Image-Text Matching~\cite{liAlignFuseVision2021}, and Masked Language Modeling~\cite{devlinBERTPretrainingDeep2019}, for multi-modal representation alignment and understanding, and one multi-modal text generation task, \textit{i.e.,} PrefixLM~\cite{biPALMPretrainingAutoencoding2020} for multi-modal text-generation, respectively. 
In this study, all the experiments are conducted with mPLUG-base model, which adopts one pre-trained with CLIP-ViT~\cite{radfordLearningTransferableVisual2021} (ViT-B/16) as visual encoder and two $BERT_{base}$~\cite{devlinBERTPretrainingDeep2019} models for textual and visual feature fusing and text decoding.

\subsection{Implementation details}
Both for captioning and VQA, the visual encoder (CLIP-ViT-B/16) takes the resized 256 $\times$ 256 image as input.
The text encoder takes a caption prompt \{\textit{a picture of}\} and a  question \{\textit{Question: \{\#question\} Answer: }\} as the input for captioning and VQA, respectively.
Then the caption and answer would be generated by the following text decoder.
For the image captioning on MSCOCO Caption, we adopt iteration-based training instead of epoch-based training~\footnote{Epoch denotes how many times the model sees the complete dataset.} since such data is widely used~\cite{liAlignFuseVision2021,liBlipBootstrappingLanguageimage2022,liMPLUGEffectiveEfficient2022} and has been learned in the pre-training.
In contrast, we used epoch-based training for VQA on VQAv2 for it is not introduced in the pre-training.
During training, we adopt the AdamW optimizer with a weight decay of 0.05.
The learning rate is warmed up to the highest learning rate in the first epoch, and decayed to the lowest learning rate following a cosine schedule.
All the experiments are optimized by cross-entropy loss and conducted on four V100-32G-SMX2 GPUs with a total batch size of 256.

\section{PEFT easily Over-fitting on MSCOCO Caption}
\begin{figure*}[t]
\begin{minipage}[b]{.48\linewidth}
  \centering
\epsfig{figure=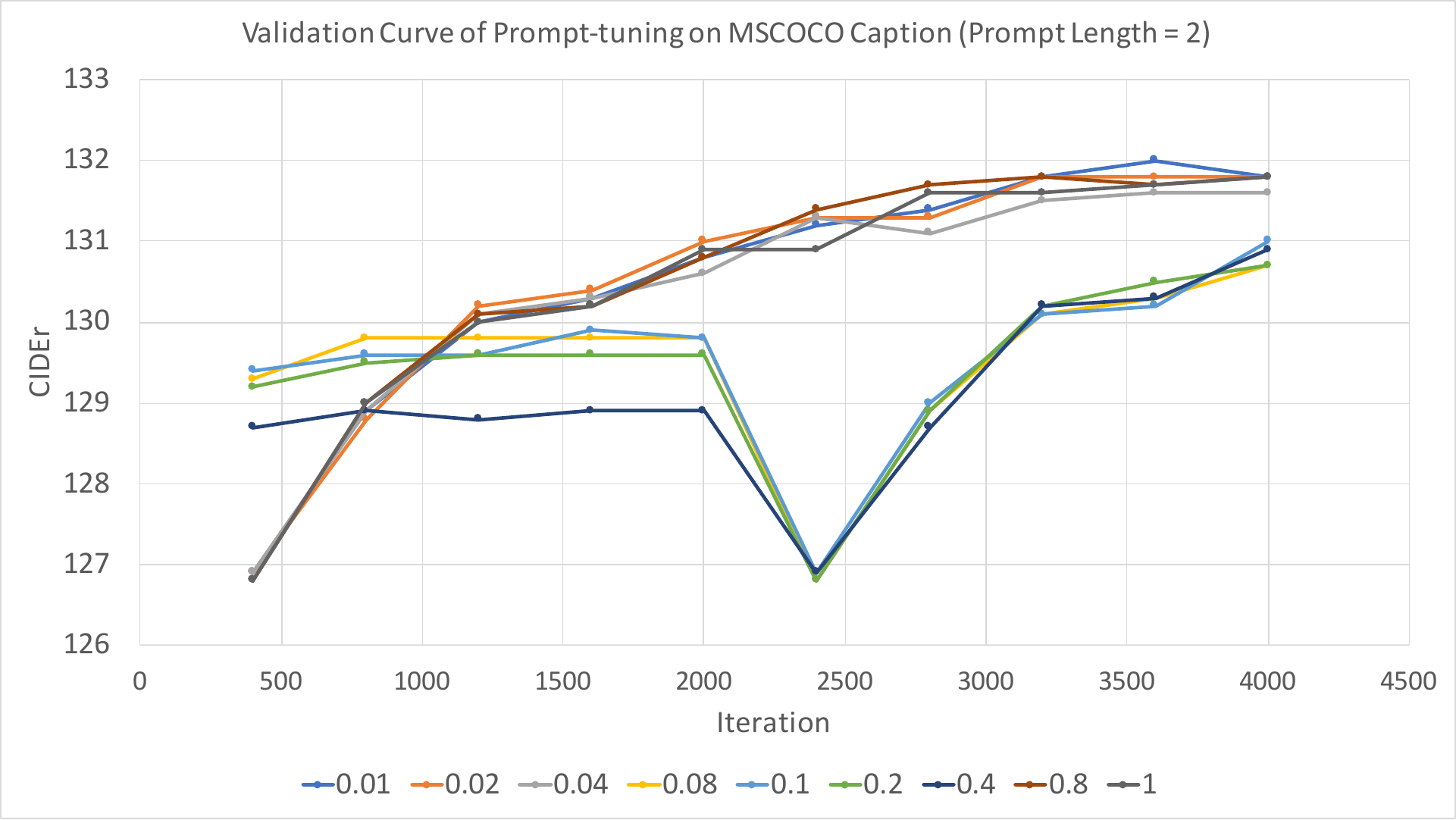,width=8.5cm}
  \vspace{1.5cm}
  \centerline{(a) Prompt-length=2.}\medskip
\end{minipage}
\hfill
\begin{minipage}[b]{0.48\linewidth}
  \centering
\epsfig{figure=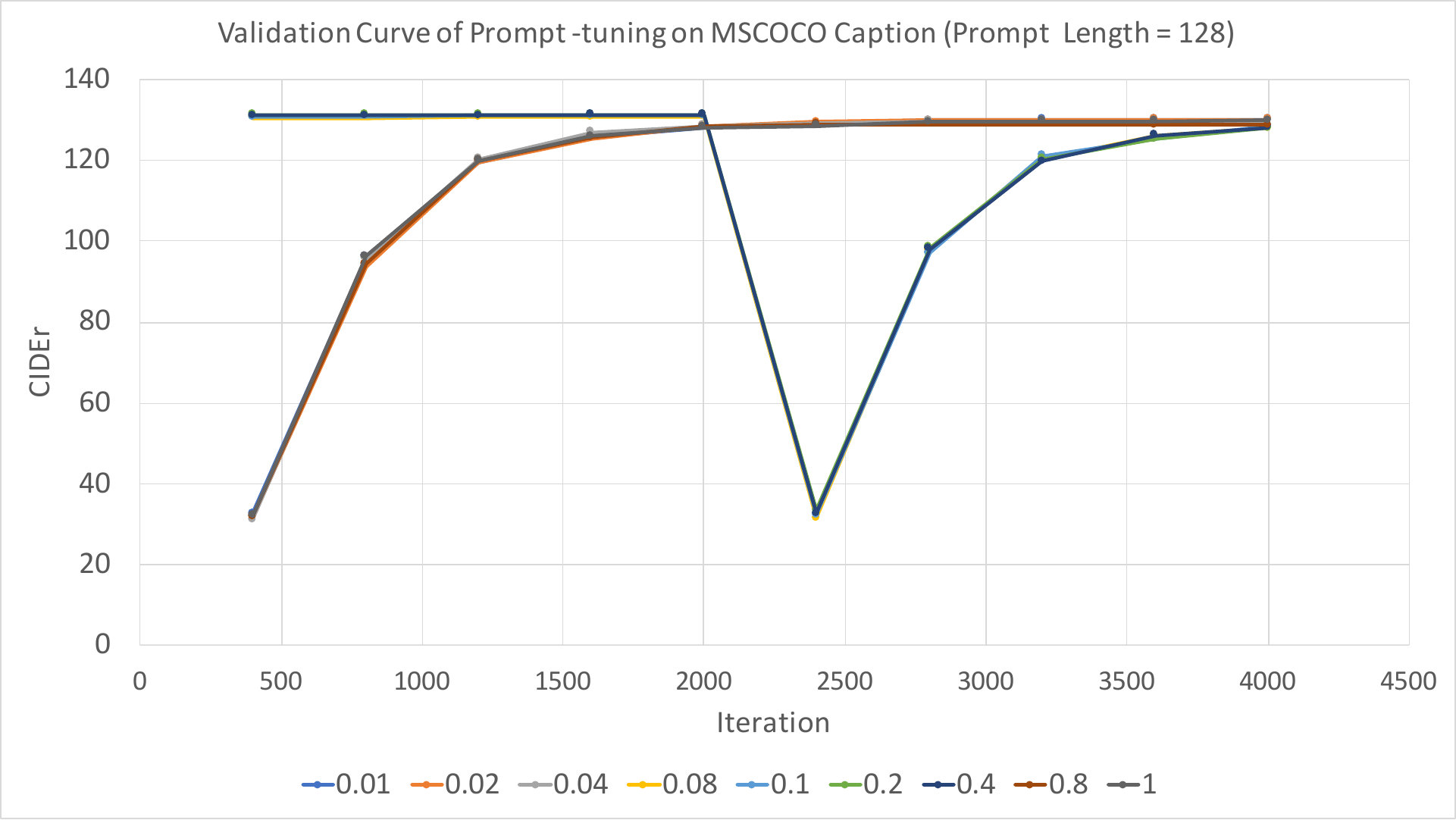,width=8.5cm}
  \vspace{1.5cm}
  \centerline{(b) Prompt-length=128.}\medskip
\end{minipage}
\caption{Validation curve of prompt-tuning on MSCOCO Caption.}
\label{fig:prompt}
\end{figure*}

\begin{figure*}[t]
\begin{minipage}[b]{.48\linewidth}
  \centering
\epsfig{figure=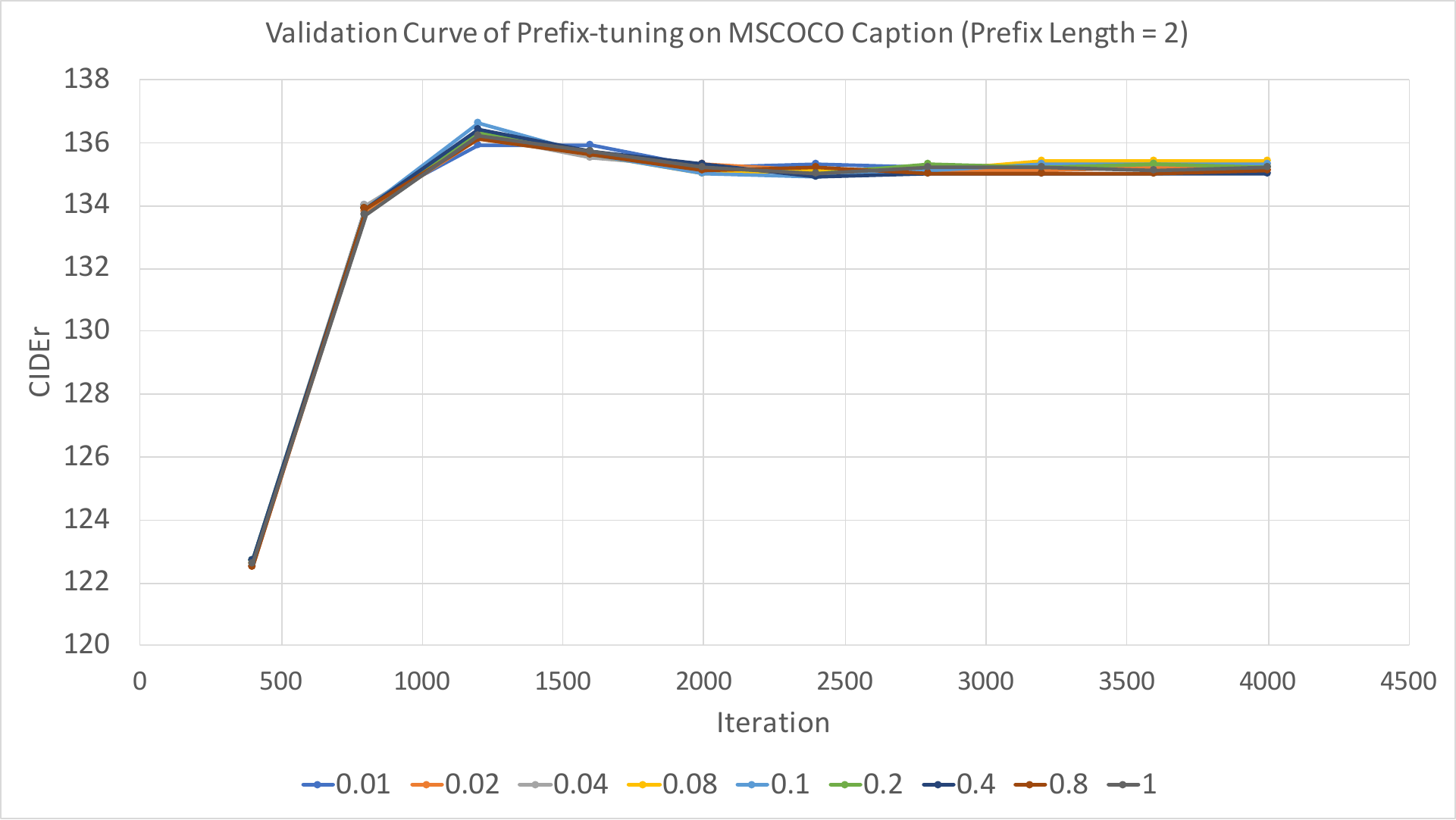,width=8.5cm}
  \vspace{1.5cm}
  \centerline{(a) Prefix-length=2.}\medskip
\end{minipage}
\hfill
\begin{minipage}[b]{0.48\linewidth}
  \centering
\epsfig{figure=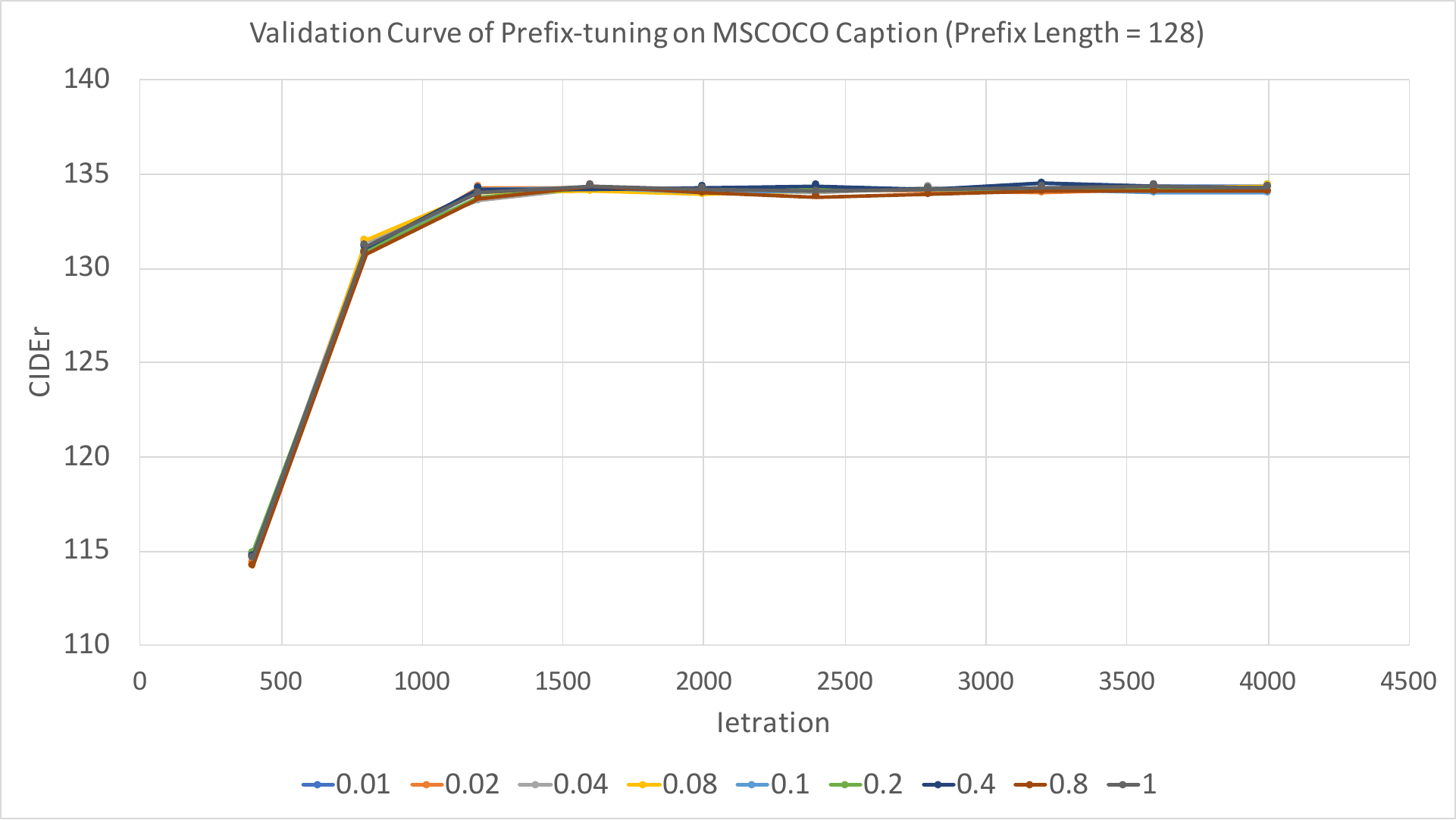,width=8.5cm}
  \vspace{1.5cm}
  \centerline{(b) Prefix-length=128.}\medskip
\end{minipage}
\caption{Validation curve of prefix-tuning on MSCOCO Caption.}
\label{fig:prefix}
\end{figure*}

\begin{figure*}[t]
\begin{minipage}[b]{.48\linewidth}
  \centering
\epsfig{figure=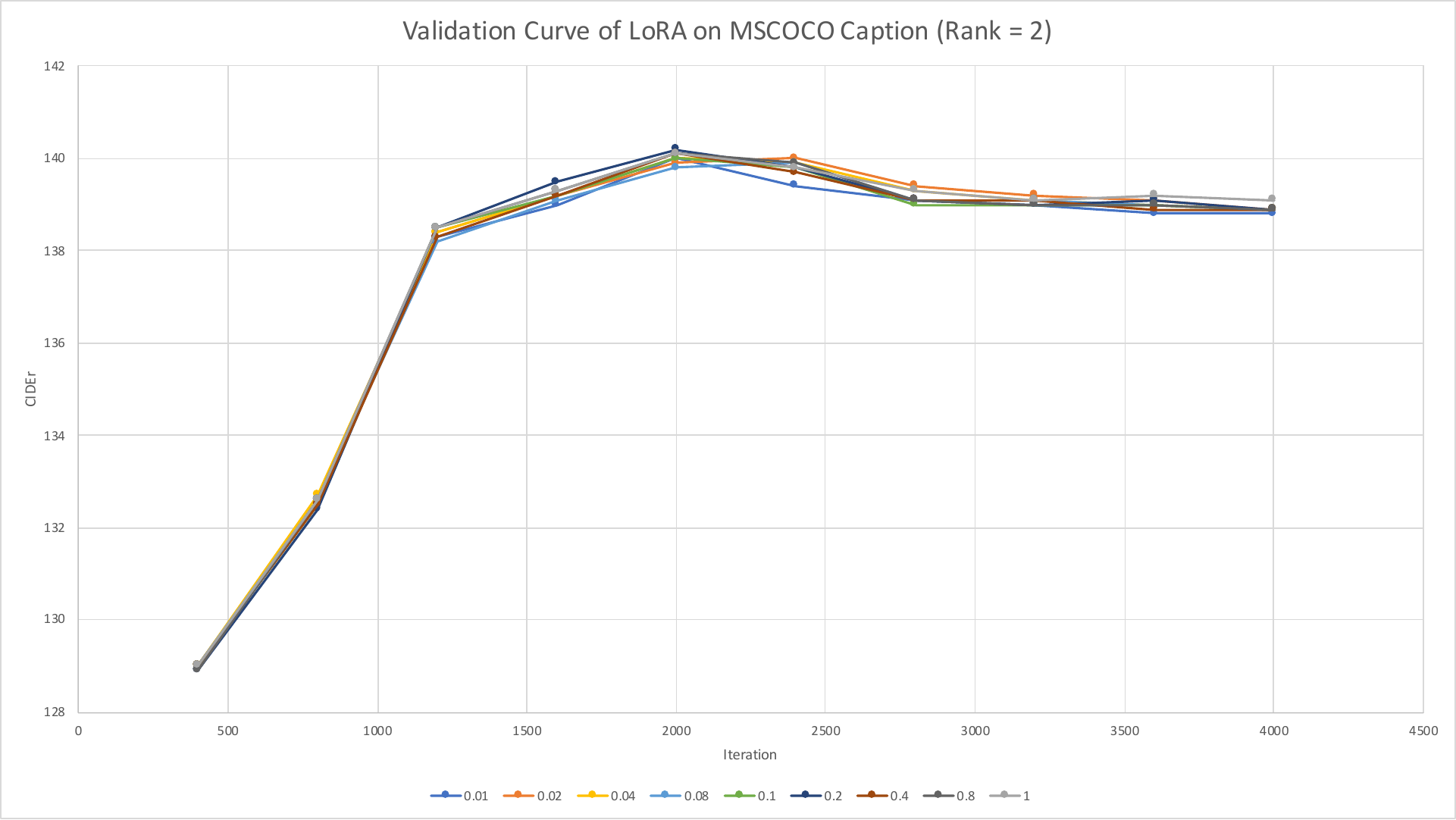,width=8.5cm}
  \vspace{1.5cm}
  \centerline{(a) Rank=2.}\medskip
\end{minipage}
\hfill
\begin{minipage}[b]{0.48\linewidth}
  \centering
\epsfig{figure=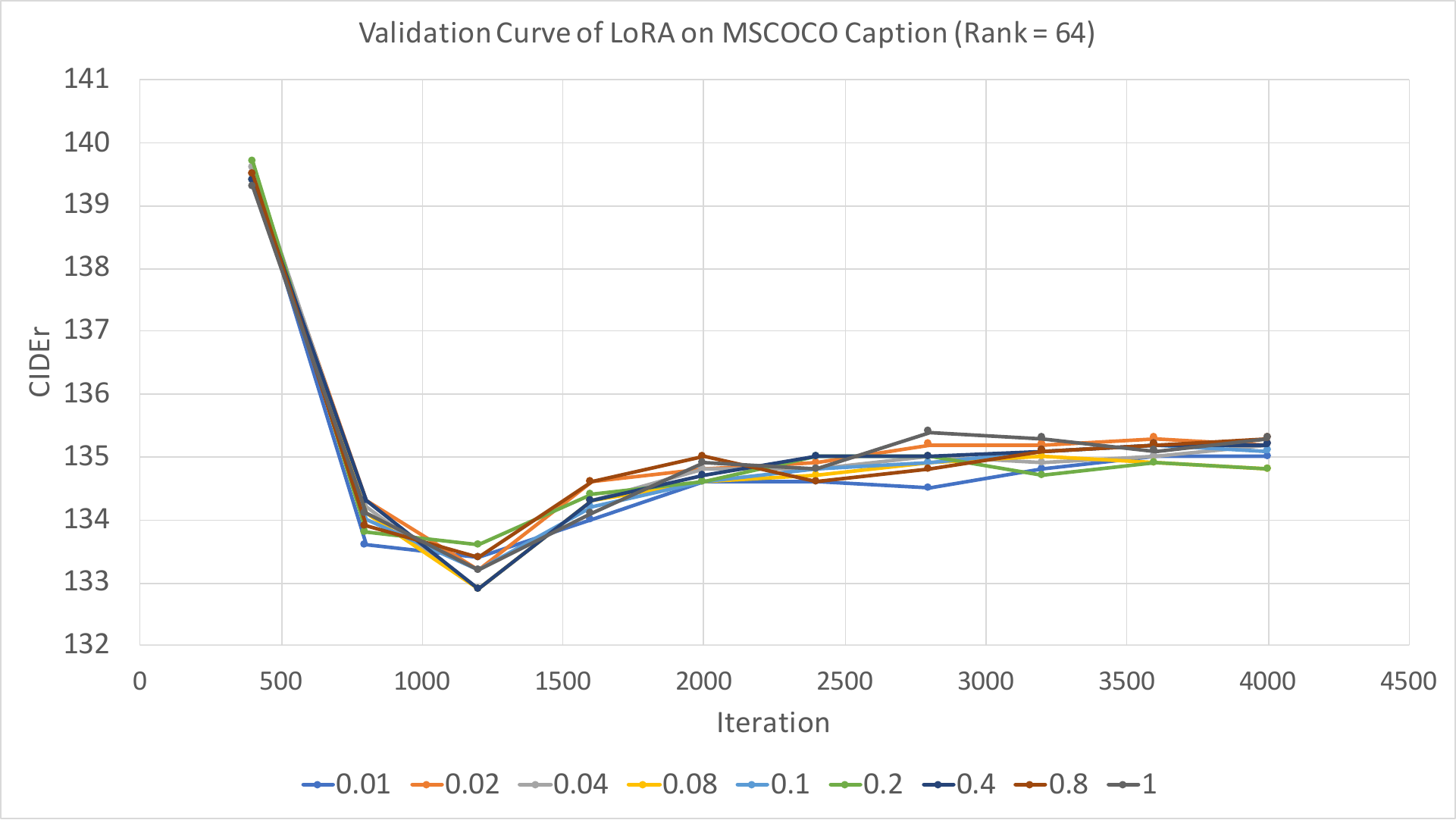,width=8.5cm}
  \vspace{1.5cm}
  \centerline{(b) Rank=128.}\medskip
\end{minipage}
\caption{Validation curve of LoRA on MSCOCO Caption.}
\label{fig:lora}
\end{figure*}

\begin{figure*}[t]
\begin{minipage}[b]{.48\linewidth}
  \centering
\epsfig{figure=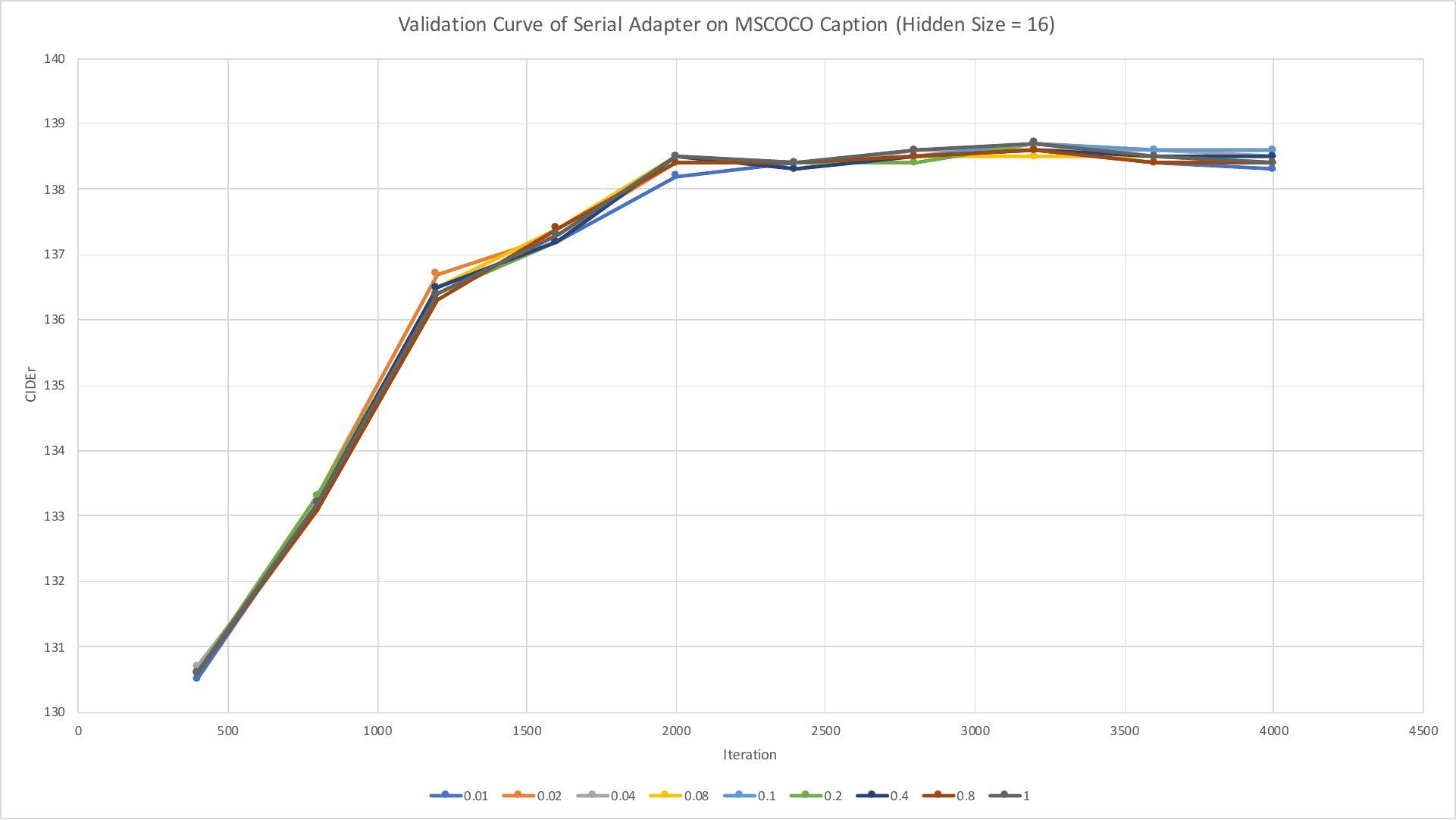,width=8.5cm}
  \vspace{1.5cm}
  \centerline{(a) Hidden-size=16.}\medskip
\end{minipage}
\hfill
\begin{minipage}[b]{0.48\linewidth}
  \centering
\epsfig{figure=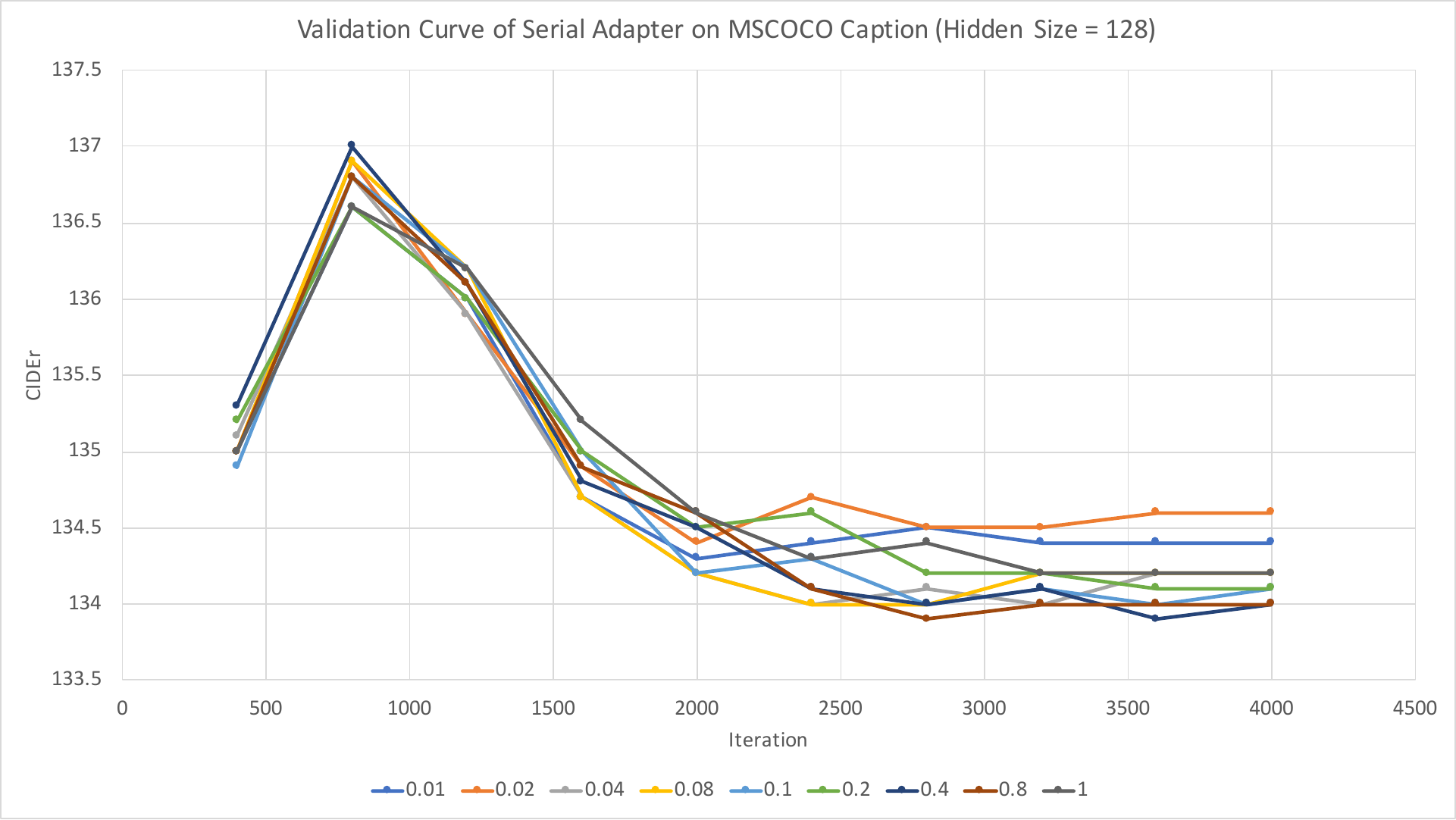,width=8.5cm}
  \vspace{1.5cm}
  \centerline{(b) Hidden-size=128.}\medskip
\end{minipage}
\caption{Validation curve of serial adapter on MSCOCO Caption.}
\label{fig:sa}
\end{figure*}

\begin{figure*}[t]
\begin{minipage}[b]{.48\linewidth}
  \centering
\epsfig{figure=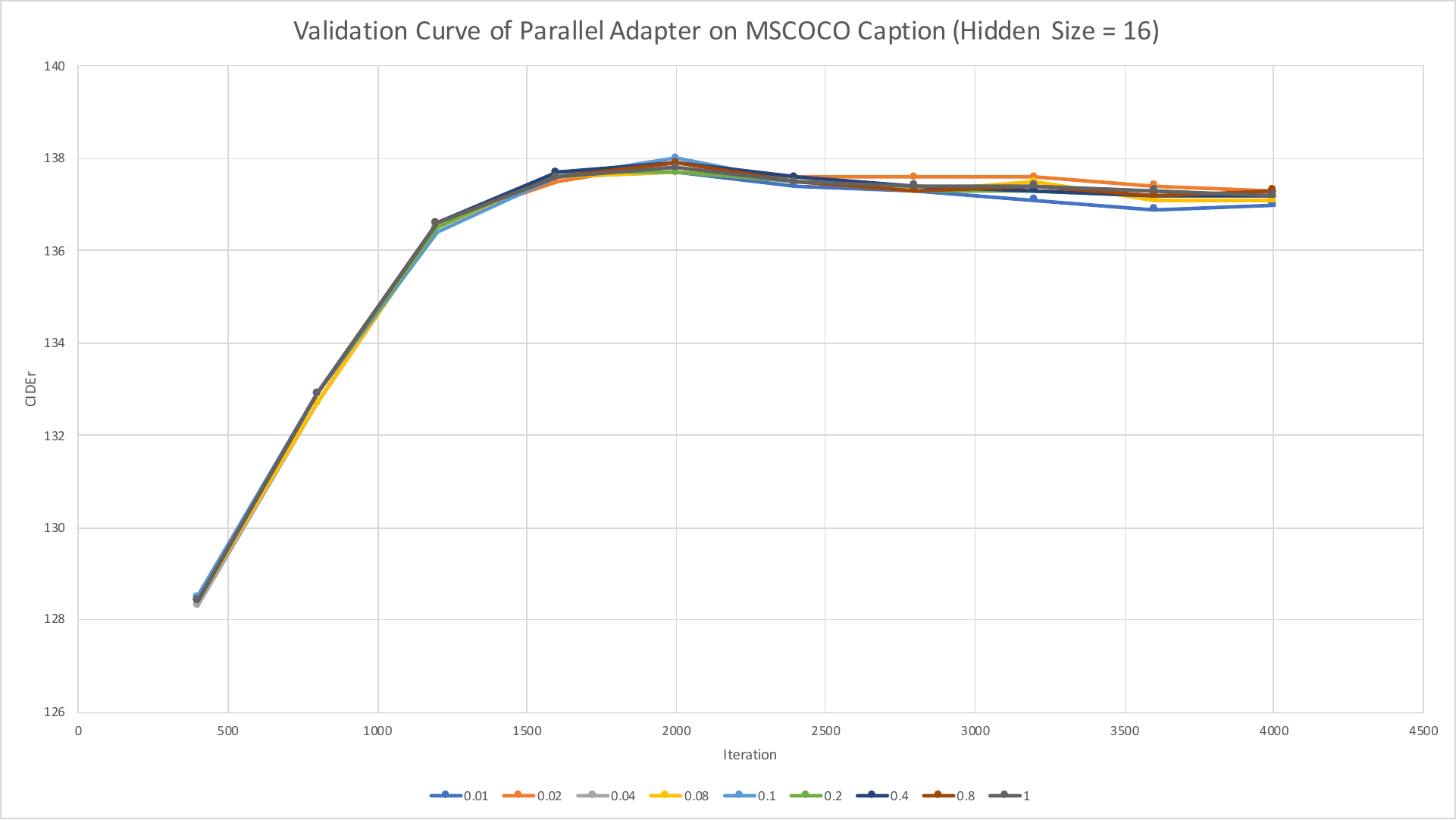,width=8.5cm}
  \vspace{1.5cm}
  \centerline{(a) Hidden-size=16.}\medskip
\end{minipage}
\hfill
\begin{minipage}[b]{0.48\linewidth}
  \centering
\epsfig{figure=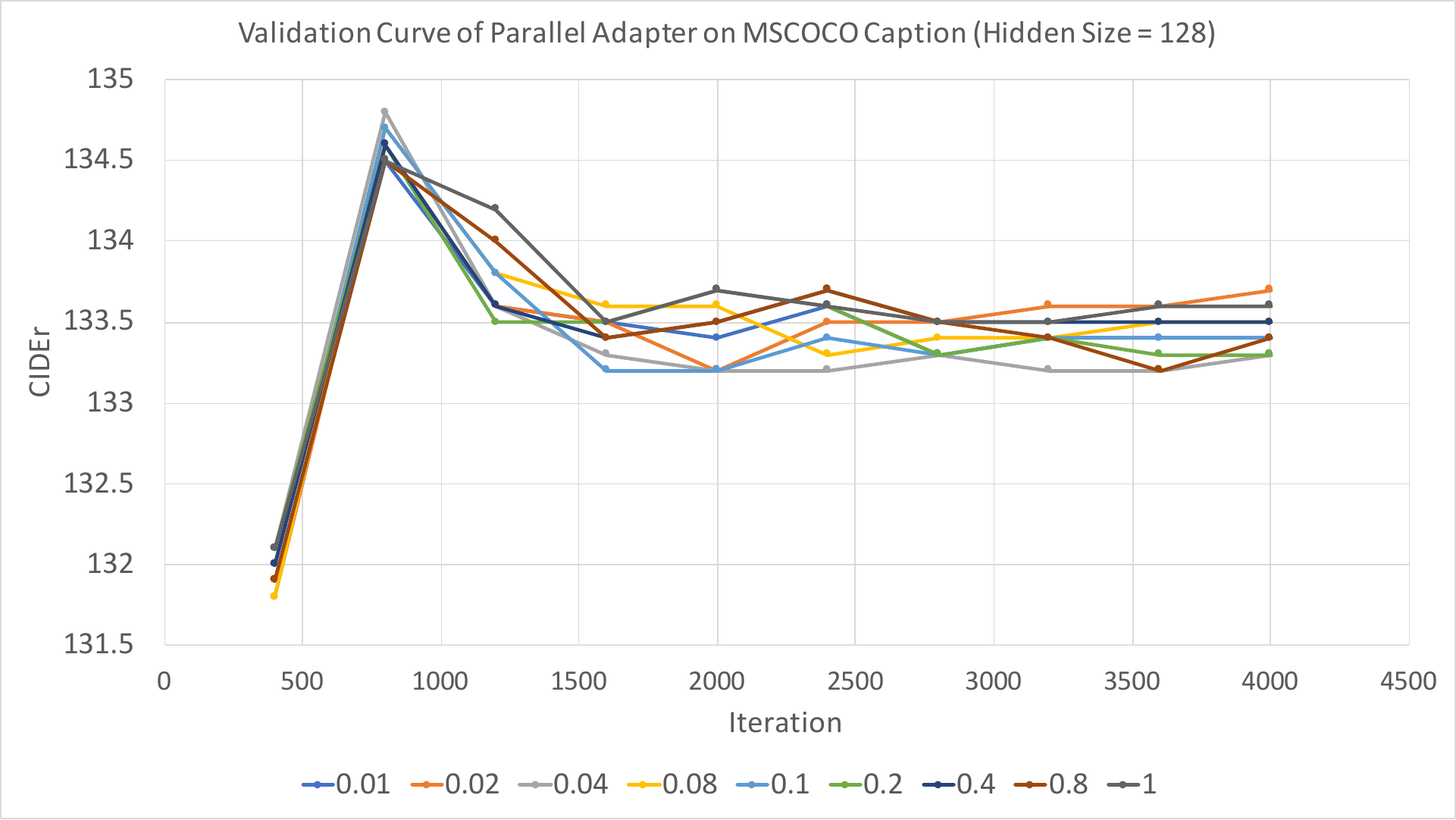,width=8.5cm}
  \vspace{1.5cm}
  \centerline{(b) Hidden-size=128.}\medskip
\end{minipage}
\caption{Validation curve of parallel adapter on MSCOCO Caption.}
\label{fig:pa}
\end{figure*}

We also provide the validation curve of the five PEFTs on MSCOCO Caption, illustrated from Fig~\ref{fig:prompt} to Fig.~\ref{fig:pa}.
One could also that since the captioning on MSCOCO Caption is consistent with the pre-training, scaling up the fine-tunable parameter size would not monotonously improve the performance of PEFTs.
We argue that if the downstream task is consistent with pre-training, too many fine-tunable parameters would lead to over-fitting.

\begin{table}[h]
\caption{Dataset Statistics}
\resizebox{\columnwidth}{!}{
\centering
\begin{tabular}{c|cc|cc}
\toprule
\textbf{Split} & \multicolumn{2}{c|}{\textbf{VQAv2}} & \multicolumn{2}{c}{\textbf{MSCOCO Caption}}\\
\midrule
  & Images & QA Pairs & Images & Captions \\
Train & 113.2K & 605.1K & 113.2K & 566.8K \\
Val & 5.0K & 26.7K & 5.0K & 5.0K \\
Test & 5.0K & 26.3K & 5.0K & 5.0K \\
\bottomrule
\end{tabular}
}
\label{tab:dataset}
\end{table}

\begin{sidewaystable*}
\caption{Implementation Details for the parameter size.}
\begin{tabular}{c|c|c|c|c|c|c}
\toprule
\textbf{Dataset}        & \textbf{Method}  & \textbf{Data Ratio}                                                                         & \textbf{Highest LR} & \textbf{LR Scheduling}                                                                                                                   & \textbf{Epochs/Steps} & \textbf{Modulated Hyper-paramter}                              \\ \midrule
\multirow{15}{*}{MSCOCO} & Prompt-tuning    & \begin{tabular}[c]{@{}c@{}}1\%, 2\%, 4\%, 8\%, \\ 10\%, 20\%,40\%, 80\%, 100\%\end{tabular} & 3e-4                & \begin{tabular}[c]{@{}c@{}}Warmup from 0 to highest LR \\ at the first 400 steps \\ and cosine decay to 0\end{tabular}                   & 4000 Steps            & \begin{tabular}[c]{@{}c@{}}The length of prompt:\\ 2,4,8,16,32,64,128\end{tabular}   \\ \cmidrule{2-7} 
                        & Prefix-tuning    & \begin{tabular}[c]{@{}c@{}}1\%, 2\%, 4\%, 8\%, \\ 10\%, 20\%,40\%, 80\%, 100\%\end{tabular} & 3e-4                & \begin{tabular}[c]{@{}c@{}}Warmup from 0 to highest LR \\ at the first 400 steps \\ and cosine decay to 0\end{tabular}                   & 4000 Steps            & \begin{tabular}[c]{@{}c@{}}The length of prefix:\\ 2,4,8,16,32,64,128\end{tabular}   \\ \cmidrule{2-7} 
                        & LoRA             & \begin{tabular}[c]{@{}c@{}}1\%, 2\%, 4\%, 8\%, \\ 10\%, 20\%,40\%, 80\%, 100\%\end{tabular} & 5e-6                & \begin{tabular}[c]{@{}c@{}}Warmup from 0 to highest LR \\ at the first 400 steps \\ and cosine decay to 0\end{tabular}                   & 4000 Steps            & \begin{tabular}[c]{@{}c@{}}The rank of LoRA:\\ 2,4,8,16,32\end{tabular}                                                     \\ \cmidrule{2-7} 
                        & Serial Adapter   & \begin{tabular}[c]{@{}c@{}}1\%, 2\%, 4\%, 8\%, \\ 10\%, 20\%,40\%, 80\%, 100\%\end{tabular} & 1e-6                & \begin{tabular}[c]{@{}c@{}}Warmup from 0 to highest LR \\ at the first 400 steps \\ and cosine decay to 0\end{tabular}                   & 4000 Steps            & \begin{tabular}[c]{@{}c@{}}The hidden size of \\ serial adapter:\\ 16,32,64,128\end{tabular}                                                    \\ \cmidrule{2-7} 
                        & Parallel Adapter & \begin{tabular}[c]{@{}c@{}}1\%, 2\%, 4\%, 8\%, \\ 10\%, 20\%,40\%, 80\%, 100\%\end{tabular} & 1e-6                & \begin{tabular}[c]{@{}c@{}}Warmup from 0 to highest LR \\ at the first 400 steps \\ and cosine decay to 0\end{tabular}                   & 4000 Steps            & \begin{tabular}[c]{@{}c@{}}The hidden size of \\ parallel adapter:\\ 16,32,64,128\end{tabular}                                                    \\ \midrule
\multirow{15}{*}{VQAv2}  & Prompt-tuning    & 10\%, 20\%,40\%, 80\%, 100\%                                                                & 3e-4                & \begin{tabular}[c]{@{}c@{}}Warmup from 1e-6 to highest LR \\ at the first 800*data\_ratio steps \\ and cosine decay to 1e-6\end{tabular} & 10 Epochs             & \begin{tabular}[c]{@{}c@{}}The length of prompt:\\ 2,4,8,16,32,64,128\end{tabular}   \\ \cmidrule{2-7} 
                        & Prefix-tuning    & 10\%, 20\%,40\%, 80\%, 100\%                                                                & 3e-4                & \begin{tabular}[c]{@{}c@{}}Warmup from 1e-6 to highest LR \\ at the first 800*data\_ratio steps \\ and cosine decay to 1e-6\end{tabular} & 10 Epochs             & \begin{tabular}[c]{@{}c@{}}The length of prefix:\\ 2,4,8,16,32,64,128\end{tabular}   \\ \cmidrule{2-7} 
                        & LoRA             & 10\%, 20\%,40\%, 80\%, 100\%                                                                & 3e-4                & \begin{tabular}[c]{@{}c@{}}Warmup from 1e-6 to highest LR \\ at the first 800*data\_ratio steps \\ and cosine decay to 1e-6\end{tabular} & 10 Epochs             & \begin{tabular}[c]{@{}c@{}}The rank of LoRA:\\ 2,4,8,16,32\end{tabular}                                                     \\ \cmidrule{2-7} 
                        & Serial Adapter   & 10\%, 20\%,40\%, 80\%, 100\%                                                                & 3e-4                & \begin{tabular}[c]{@{}c@{}}Warmup from 1e-6 to highest LR \\ at the first 800*data\_ratio steps \\ and cosine decay to 1e-6\end{tabular} & 10 Epochs             & \begin{tabular}[c]{@{}c@{}}The hidden size of \\ serial adapter\\ 16,32,64,128,256,512\end{tabular} \\ \cmidrule{2-7} 
                        & Parallel Adapter & 10\%, 20\%,40\%, 80\%, 100\%                                                                & 3e-4                & \begin{tabular}[c]{@{}c@{}}Warmup from 1e-6 to highest LR \\ at the first 800*data\_ratio steps \\ and cosine decay to 1e-6\end{tabular} & 10 Epochs             & \begin{tabular}[c]{@{}c@{}}The hidden size of \\ parallel adapter\\ 16,32,64,128,256,512\end{tabular} \\ \bottomrule
\end{tabular}
\end{sidewaystable*}

\clearpage

\end{document}